\documentclass[11pt]{article}

\usepackage[preprint]{acl}

\usepackage{times}
\usepackage{latexsym}

\usepackage[T1]{fontenc}
\usepackage[utf8]{inputenc}

\usepackage{microtype}

\usepackage{inconsolata}

\usepackage{graphicx}
\usepackage[most]{tcolorbox}
\usepackage{makecell}
\usepackage{booktabs}
\usepackage{multirow}
\usepackage[table]{xcolor} % for \cellcolor inside tables
\definecolor{Better}{RGB}{198,239,206} % light green
\definecolor{Worse}{RGB}{255,199,206}  % light red
\definecolor{bluebg}{RGB}{222,235,247}   % 淺藍底
\definecolor{greenbg}{RGB}{226,239,217}  % 淺綠底
\definecolor{orangebg}{RGB}{255,235,205} % 淺橘底

\usepackage{xcolor}
\usepackage{listings}
\usepackage{amsfonts}
\usepackage[most]{tcolorbox}
\usepackage{caption} % for \caption in figure*

\usepackage{xcolor}
\usepackage{listings}
\usepackage[most]{tcolorbox}
\usepackage{caption} % for \captionof
\tcbuselibrary{listings, breakable}

\lstdefinestyle{promptmono}{
  language=Python,
  basicstyle=\ttfamily\footnotesize,
  columns=fullflexible,
  breaklines=true,
  showstringspaces=false,
  tabsize=2,
  upquote=true
}

\tcbset{
  emnlpPrompt/.style={
    colback=gray!3,
    colframe=gray!35,
    coltitle=black,
    fonttitle=\bfseries\footnotesize,
    title filled,
    boxrule=1.0pt,
    arc=1pt,
    left=6pt,right=6pt,top=6pt,bottom=6pt,
    listing only,
    listing options={style=promptmono}
  }
}
\tcbset{
  mybox/.style={
    colback=blue!2!white,   % light background
    colframe=blue!30!black, % border color
    boxrule=0.6pt,          % border thickness
    arc=3pt,                % rounded corners
    left=6pt, right=6pt,
    top=4pt, bottom=4pt,
    boxsep=0pt,
    enhanced,
    fonttitle=\bfseries,
    coltitle=black,
  }
}
\usepackage{tcolorbox}
\tcbuselibrary{listings, skins, breakable}

\tcbset{
  promptbox/.style={
    colback=blue!2!white,   % light background
    colframe=blue!30!black,
    fonttitle=\bfseries,
    breakable,
    boxrule=0.5pt,
    left=3mm,
    right=3mm,
    top=2mm,
    bottom=2mm,
    before skip=2em,   % space before each box
    after skip=2em     % space after each box
  }
}

\title{Beyond Facts: Benchmarking Distributional Reading Comprehension in Large Language Models}

\author{
  \textbf{Pei-Fu Guo\textsuperscript{1}},
  \textbf{Ya-An Tsai\textsuperscript{1}},
  \textbf{Chun-Chia Hsu\textsuperscript{1}},
  \textbf{Kai-Xin Chen\textsuperscript{1}},
  \textbf{Yun-Da Tsai\textsuperscript{1}},
\\
  \textbf{Kai-Wei Chang\textsuperscript{2}},
  \textbf{Nanyun Peng\textsuperscript{2}},
  \textbf{Mi-Yen Yeh\textsuperscript{3}},
  \textbf{Shou-De Lin\textsuperscript{1,4}}
\\
\\
  \textsuperscript{1}National Taiwan University
  \textsuperscript{2}University of California, Los Angeles
\\
  \textsuperscript{3}Academia Sinica, Taiwan
  \textsuperscript{4}NTU AI-CoRE
\\
  \small{
    \textbf{Correspondence:} \href{mailto:r12922217@csie.ntu.edu.tw}{r12922217@csie.ntu.edu.tw}
  }
}

\begin{document}
\maketitle

\begin{abstract}
% Large language models (LLMs) are widely used to comprehend diverse information from text. 
While most reading comprehension benchmarks for LLMs focus on factual information that can be answered by localizing specific textual evidence, many real-world tasks require understanding distributional information, such as population-level trends and preferences expressed across collections of text.
We introduce \textsc{Text2DistBench}, a reading comprehension benchmark for evaluating LLMs’ ability to infer distributional knowledge from natural language. 
Built from real-world YouTube comments about movie and music entities, the benchmark provides models with entity metadata and associated comments, and requires them to answer distributional questions, such as estimating the proportions of positive and negative comments, or identifying the most and second most frequent topics discussed among viewers.
% , and joint distributions (e.g., co-occurring sentiment--topic pairs).
% marginal, conditional, and joint distributions.
To support reliable and long-term evaluation, the construction pipeline of \textsc{Text2DistBench} is fully automated and continuously updated to incorporate newly emerging entities over time.
Experiments across multiple LLMs show that while models substantially outperform random baselines, performance varies widely across different distribution types and characteristics.
These findings highlight both the capabilities and limitations of current LLMs in distributional reading comprehension and demonstrate the value of \textsc{Text2DistBench} as a practical and scalable testbed for future research.\footnote{Code and data are available at \href{https://github.com/0Frett/Text2DistBench}{link}.}
\end{abstract}

\section{Introduction}
\label{sec:intro}

% Large language models (LLMs) are increasingly used as general-purpose language understanding systems, where they must comprehend diverse information from text and produce appropriate responses. Most existing reading comprehension benchmarks focus on evaluate information that is generally
% \emph{explicitly stated} in text, where the correct answer can be directly retrieved from a specific span~\citep{rajpurkar2016squad100000questionsmachine,tacl_a_00276,dua2019dropreadingcomprehensionbenchmark} or a small set of localized evidence~\citep{tafjord-etal-2019-quartz, welbl-etal-2018-constructing}. 
Large language models (LLMs) are increasingly used as general-purpose language understanding systems, where they must comprehend diverse information from text and produce appropriate responses. Most existing reading comprehension benchmarks focus on \emph{factual knowledge}, where answers can be obtained by retrieving or reasoning over specific pieces of textual evidence, such as sentence spans~\citep{rajpurkar2016squad100000questionsmachine,tacl_a_00276,dua2019dropreadingcomprehensionbenchmark} or a set of localized facts~\citep{tafjord-etal-2019-quartz, welbl-etal-2018-constructing}.

\begin{figure}
    \centering
    \includegraphics[width=\linewidth]{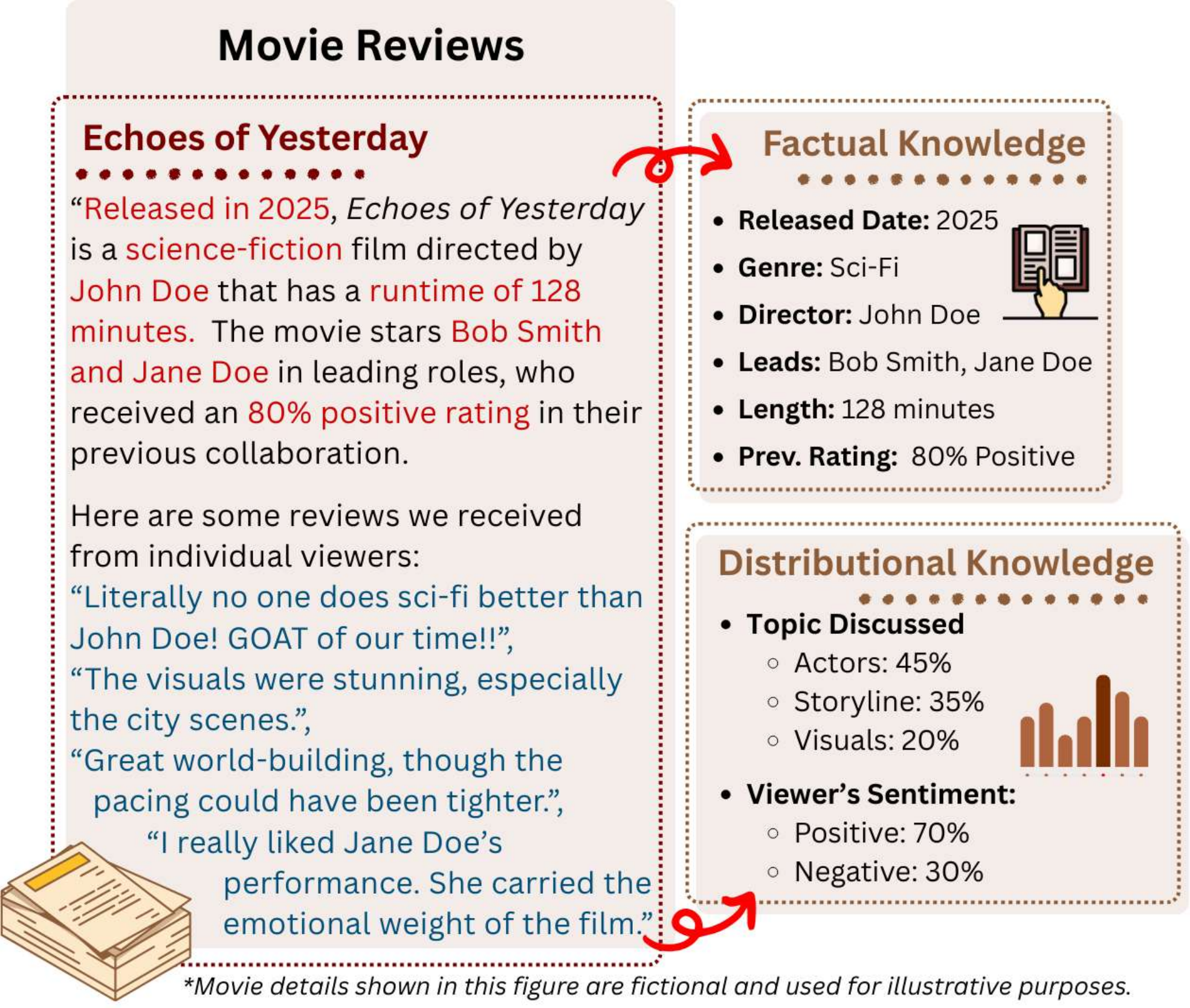}
    \caption{\textbf{Factual vs.\ Distributional Knowledge.}}
    \label{fig:motivation}
\end{figure}

However, many real-world information needs go beyond factual knowledge and require models to understand \emph{distributional knowledge} expressed across text.
As illustrated in Figure~\ref{fig:motivation}, factual information such as movie genre or director is localized and can be answered by identifying specific textual evidence.
In contrast, distributional information, such as proportion of viewer sentiment or topic prevalence, must be derived by aggregating patterns across many individual comments.
This form of understanding is important as LLMs are increasingly used to support human decision making and analyze public opinion in real-world settings such as market research and product review analysis, where users summarize collective opinions, identify dominant trends, and compare preferences across aspects from large volumes of text.

Recent work has explored models’ \emph{distributional knowledge} from different perspectives.
Some benchmarks study models' distributional knowledge of human values, using survey-style questions to probe models' beliefs about cultural values or political attitudes associated with specific demographic groups~\citep{durmus2023measuring,meister2025benchmarking,röttger2024politicalcompassspinningarrow,zhao2024worldvaluesbenchlargescalebenchmarkdataset}.
Other benchmarks focus on probabilistic reasoning, where models are provided with explicitly specified distributions (e.g., Poisson distributions) or numeric data and are asked to calculate statistics or sample outcomes~\citep{paruchuri2024oddslanguagemodelscapable,pournemat2025reasoninguncertaintyexploringprobabilistic,ozturkler2023thinksumprobabilisticreasoningsets}.
While valuable, these benchmarks primarily probe distributional knowledge encoded in models during pretraining.
They do not evaluate whether models can derive population-level distributions, such as the \emph{frequencies} or \emph{prevalences} of different opinions, by reading and aggregating information from natural language text.

% This leaves distributional understanding grounded in text underexplored.

% While valuable, these benchmarks do not evaluate whether models can derive population-level distributions directly from natural language text, leaving an important gap in the evaluation of distributional understanding.

% The benchmark is constructed from real-world movie and music user comments collected from YouTube, and is designed to probe how models infer population-level distributions from natural language content.

Based on this gap, we introduce \textsc{Text2Dist-}\textsc{Bench}, a reading comprehension benchmark designed to systematically evaluate LLMs’ ability to infer \emph{distributional knowledge} from text.
Constructed from real-world YouTube comments, the benchmark provides models with metadata and user comments for 
% previously unseen 
movie and music entities released after the model knowledge cutoff date. 
% Each benchmark question requires models to comprehend and aggregate information across the provided content to infer population-level distributions.
Specifically, each benchmark instance includes:
(1) \textbf{entity metadata}, which provides background information needed to interpret the comments;
(2) a set of \textbf{human comments} associated with the entity; and
(3) \textbf{distributional QA pairs}, which require models to aggregate information across comments to infer population-level statistics (e.g., estimating the percentage of positive and negative comments or identifying the most frequently discussed topics).

To define the underlying distributions, each comment is automatically annotated using multiple LLMs across two attributes: \emph{sentiment} (e.g., positive or negative) and \emph{topic} (e.g., acting, storyline, visuals, or audio). These annotations induce a discrete distribution over attributes for each entity.
Based on this formulation, we design questions that probe different aspects of distributional understanding. Specifically, models are asked to estimate category proportions (\emph{estimation}) and to identify the most and second most frequent categories (\emph{mode queries}). These questions are instantiated over marginal distributions (e.g., overall sentiment or topic prevalence), conditional distributions (e.g., viewer sentiment given a topic), and joint distributions that capture attribute co-occurrence patterns (e.g. percentage of users that are positive on the lyrics of a song).

In addition, \textsc{Text2DistBench} is constructed through a fully automated pipeline that supports continuous updates. By continuously introducing newly emerging entities, the benchmark reduces the risk of data leakage, where models may have already been exposed to the entity-related information during training, and instead encourages them to derive distributions from the provided text. This design makes \textsc{Text2DistBench} more sustainable and allows it to remain reliable as LLMs and their training data evolve over time.

Using \textsc{Text2DistBench}, we evaluate a range of state-of-the-art LLMs. Our experiments show that while current models substantially outperform random baseline, their performance varies widely across distributional settings. In particular, models generally perform better on questions related to marginal distributions than on conditional or joint distributions, and are sensitive to intrinsic properties of the target distribution (e.g., uniformity, probability mass concentration).
Moreover, we find that models can form informative prior beliefs from factual information alone, which often closely approximate the target distributions even for unseen entities.
These findings highlight both the strengths and limitations of current LLMs in understanding population-level information from text and demonstrate the utility of \textsc{Text2DistBench} for distributional reading comprehension.

\section{Related Work}
\label{sec:related_work}

\paragraph{Reading Comprehension Benchmarks}
Most reading comprehension benchmarks focus on evaluating \emph{factual knowledge} that can be derived from text, where answers can be obtained by retrieving or reasoning over specific pieces of textual evidence, such as sentence spans or localized facts.
Representative benchmarks include SQuAD, WikiReading, Natural Questions, and DROP, which evaluate factual understanding over short or long documents \citep{rajpurkar2016squad100000questionsmachine,hewlett-etal-2016-wikireading,tacl_a_00276,dua2019dropreadingcomprehensionbenchmark}.
Other benchmarks emphasize more advanced forms of comprehension, such as contextual commonsense reasoning~\citep{huang-etal-2019-cosmos}, qualitative relations~\citep{tafjord-etal-2019-quartz}, and multi-hop reasoning~\citep{welbl-etal-2018-constructing}.
In contrast, \textsc{Text2DistBench} focuses on distributional knowledge, where the goal is to infer population-level patterns and statistics by aggregating information across a collection of textual comments.

\paragraph{Probabilistic Reasoning Benchmarks}
A separate line of work evaluates models’ understanding of probabilistic concepts~\citep{paruchuri2024oddslanguagemodelscapable,pournemat2025reasoninguncertaintyexploringprobabilistic,freedman2025exploringpotentiallargelanguage}. 
In these benchmarks, models are typically provided with explicitly specified distributions (e.g., Normal or Poisson distributions) or tabular numerical data, and are asked to estimate percentiles, sample outcomes, or compute specific probabilities.
In contrast to these settings, where the underlying distributions are directly given or numerically represented, \textsc{Text2DistBench} requires models to infer distributional properties from unstructured natural language.

% modified
% Unlike these benchmarks, where the underlying distributions are formally
% specified, \textsc{Text2DistBench} requires models to infer distributional
% properties from unstructured natural language, without access to explicit
% mathematical definitions.

\paragraph{Human Values Distribution Benchmarks}
Recent work has evaluated LLMs by comparing their response distributions to
human distributions, often using survey-style questions and divergence-based
metrics~
\citep{durmus2023measuring,meister2025benchmarking}.
These benchmarks typically probe models’ beliefs about cultural values~\citep{Naous2024,Wang2024}, political attitudes~\citep{röttger2024politicalcompassspinningarrow,Stammbach2024}, and public opinions~\citep{zhao2024worldvaluesbenchlargescalebenchmarkdataset} associated with specific demographic groups.
% In contrast, \textsc{Text2DistBench} evaluates distributional understanding
% conditioned on text, requiring models to derive population-level statistics from the provided context rather than relying on pretrained knowledge.
In contrast, \textsc{Text2DistBench} evaluates distributional understanding grounded in text, requiring models to derive population-level statistics by aggregating information from the provided context rather than relying on pretrained knowledge.

\section{Methodology: Benchmark Overview}
\label{sec:dataset}

\textsc{\textsc{Text2DistBench}} is a reading comprehension benchmark designed to evaluate a model’s ability to understand \emph{distributional knowledge} from text. 
% We use movie and music reviews as data sources and construct reading comprehension questions that probe different properties of the underlying distributions within content. 
We use movie and music comments from YouTube as data sources and construct reading comprehension questions that require models to infer different properties of the distributions expressed across the comments.
Each \textsc{\textsc{Text2DistBench}} instance consists of three components:
(1) \textit{entity metadata}, which provides contextual background for interpreting the comments;
(2) a set of \textit{human comments} associated with the entity; and
(3) \textit{distributional QA pairs}, which ask information about the underlying distribution expressed in the comments.
To enable systematic evaluation, questions are organized along two
axes: \emph{distribution type}, which specifies the distribution being queried, and \emph{task type}, which specifies how the model is asked about that distribution.
% The Cartesian product of three distribution types and three task types allow us to analyze model behavior across different forms of distributional reasoning.

\begin{table*}[t]
\centering
\small
\setlength{\tabcolsep}{6pt}
\renewcommand{\arraystretch}{1.2}
\begin{tabular}{p{0.1\linewidth}|p{0.8\linewidth}}
\toprule
\textbf{Distribution}  & \textbf{Question Example (Task Type: Most Frequent)} \\
\midrule

{Marginals} &
\begin{minipage}[t]{\linewidth}\ttfamily
$P(S)$: What overall attitude do most viewers express? Positive or Negative.\\
$P(T)$: Which aspect of the movie is discussed most often? Actor, Storyline, Visual, or Audio.
\end{minipage}\\
\midrule

Conditionals  &
\begin{minipage}[t]{\linewidth}\ttfamily
$P(S\mid T)$: Among the comments that talk about the actor, what attitude do viewers express most commonly? Positive or Negative.\\
$P(T \mid S)$: Among the comments that express a positive attitude, which aspect of the movie is mentioned most often? Actor, Storyline, Visual, or Audio.
\end{minipage}\\
\midrule

Joint  &
\begin{minipage}[t]{\linewidth}\ttfamily
$P(S, T)$: Considering both (1) which aspect (Actor, Storyline, Visual, Audio) is being talked about and (2) whether the attitude is positive or negative.
Which (aspect, sentiment) combination appears most often in the comments?
\end{minipage}\\

\bottomrule
\end{tabular}
\caption{\textbf{Question Examples of different distribution type.}}
\label{tab:knowledge_type}
\end{table*}

\begin{table*}[t]
\centering
\small
\setlength{\tabcolsep}{6pt}
\renewcommand{\arraystretch}{1.2}
\begin{tabular}{p{0.1\linewidth}|p{0.8\linewidth}}
\toprule
\textbf{Question} & \textbf{Question Example (Distribution Type: Marginals)} \\
\midrule

{Estimation} &
\begin{minipage}[t]{\linewidth}\ttfamily
Movie Information:\textcolor{blue}{\{meta\_data\}} \\YouTube Viewer Comments: \textcolor{blue}{\{comments\}}\\
What aspects of the movie do viewers talk about?\\
Summarize how frequently each aspect appears in the comments using percentages. \\Output your answer in the following format:\\
\{
    "Actor":"<int>\%",
    "Storyline":"<int>\%",
    "Visual":"<int>\%",
    "Audio":"<int>\%"
\}
\end{minipage}\\
\midrule

\makecell[tl]{Most\\Frequent}  &
\begin{minipage}[t]{\linewidth}\ttfamily
Movie Information:\textcolor{blue}{\{meta\_data\}} \\YouTube Viewer Comments: \textcolor{blue}{\{comments\}}\\
Which aspect of the movie is discussed most often?\\
Actor, Storyline, Visual, or Audio.
\end{minipage}\\
\midrule

\makecell[tl]{Second\\Frequent}  &
\begin{minipage}[t]{\linewidth}\ttfamily
Movie Information:\textcolor{blue}{\{meta\_data\}} \\YouTube Viewer Comments: \textcolor{blue}{\{comments\}}\\
Which aspect of the movie is discussed the second most often?\\
Actor, Storyline, Visual, or Audio.
\end{minipage}\\

\bottomrule
\end{tabular}
\caption{\textbf{Question Examples of different task type}}
\label{tab:task_type}
\end{table*}

\subsection{Distribution Type}
\label{sec:dataset:knowledge}
In real-world scenarios, distributional information can come from different types of distributions. For example, in product reviews that include ratings and topic tags, determining whether overall feedback is mostly positive or negative corresponds to a marginal distribution, while analyzing sentiment toward a specific product feature corresponds to a conditional distribution.

In \textsc{Text2DistBench}, each viewer comment is annotated along two dimensions:
\textit{sentiment} $S$ (e.g., positive or negative) and \textit{topic} $T$ (e.g., Actor, Storyline, Visual, Audio). 
Together, these annotations define a discrete joint distribution $P(S,T)$ over the comment set.
Based on this formulation, we define three \emph{distribution types}, each corresponding to a distribution derived from $P(S,T)$:
\textbf{Marginal distributions}, such as $P(S)$ and $P(T)$, capture the overall distribution of sentiment or topic across all comments;
\textbf{Conditional distributions}, such as $P(S\mid T)$ and $P(T\mid S)$, describe how one variable is distributed given a specific value of the other;
and \textbf{Joint distributions}, $P(S,T)$, represent the full co-occurrence between sentiment and topic.
These distribution types specify which distribution a question targets.
Question examples associated with each distribution type are shown in Table~\ref{tab:knowledge_type}.

\subsection{Task Types}
\label{sec:dataset:question_types}
We design three tasks to probe model understanding for each distribution type. Question examples of each task are shown in Table~\ref{tab:task_type}.

\paragraph{(1) Estimation}
The model is asked to estimate the full numerical distribution by predicting the percentage associated with each support category.
This task evaluates whether the model can recover the overall shape of the distribution and correctly allocate probability mass across categories.

\paragraph{(2) Most Frequent}
The model is asked to identify the support category with the largest probability mass.
This task tests whether the model can reliably recognize the dominant trend in the distribution, corresponding to use cases where users primarily care about the majority outcome.

\paragraph{(3) Second Frequent}
The model is asked to identify the support category with the second-largest probability mass.
This task probes a more subtle form of distributional understanding, requiring the model to correctly rank alternatives beyond the dominant mode.

\subsection{Evaluation Metrics}
\label{sec:dataset:eval_metric}
Evaluation metrics are defined by tasks, independent of the distribution type.

\paragraph{(1) Estimation}
% modified
In this task, the model outputs a distribution.
To quantify the discrepancy between the predicted distribution $\hat{p}_i$ and
the ground-truth distribution $p_i$, we follow prior work on measuring distributional
alignment~\citep{pournemat2025reasoninguncertaintyexploringprobabilistic,meister2025benchmarking,gupta2025coinflipsmakellms} and adopt the \emph{Total Variation Distance (TVD)} as the evaluation metric.
% adopt \emph{Total Variation Distance (TVD)}.
TVD measures the minimum amount of probability mass that must be redistributed to transform $\hat{p}_i$ into $p_i$ and is invariant to the size of the support:
\begin{equation}
    \label{eq:tvd}
    \mathrm{TVD}(\hat{p}_i, p_i)
    \;=\;
    \frac{1}{2} \sum_{k=1}^{K} \left| \hat{p}_{i,k} - p_{i,k} \right|
\end{equation}
where $K$ denotes the number of support categories.
Since $\mathrm{TVD}$ is bounded in $[0,1]$, with lower values indicating better alignment, we report $1-\mathrm{TVD}$ for consistency with other metrics where higher values represent better performance.

% modified
\paragraph{(2) Most / Second Frequent}
In this task, the model identifies the category with the largest or second-largest probability mass, and performance is evaluated using classification accuracy : 
% $\mathrm{ACC} = \mathbf{1}\!\left[\hat{y}_i = y_i\right]$
% , where \(\hat{y}_i\) is the model’s prediction and \(y_i\) is the ground-truth.
\begin{equation}
\mathrm{ACC}
=
\mathbf{1}\!\left[
\hat{y}_i = y_i
\right]
\end{equation}
where \(\hat{y}_i\) is the model’s prediction and \(y_i\) is the ground-truth category label.

\section{Methodology: Benchmark Construction}
\label{sec:dataset:construction}
\begin{figure*}[t]
    \centering
    \includegraphics[width=\linewidth]{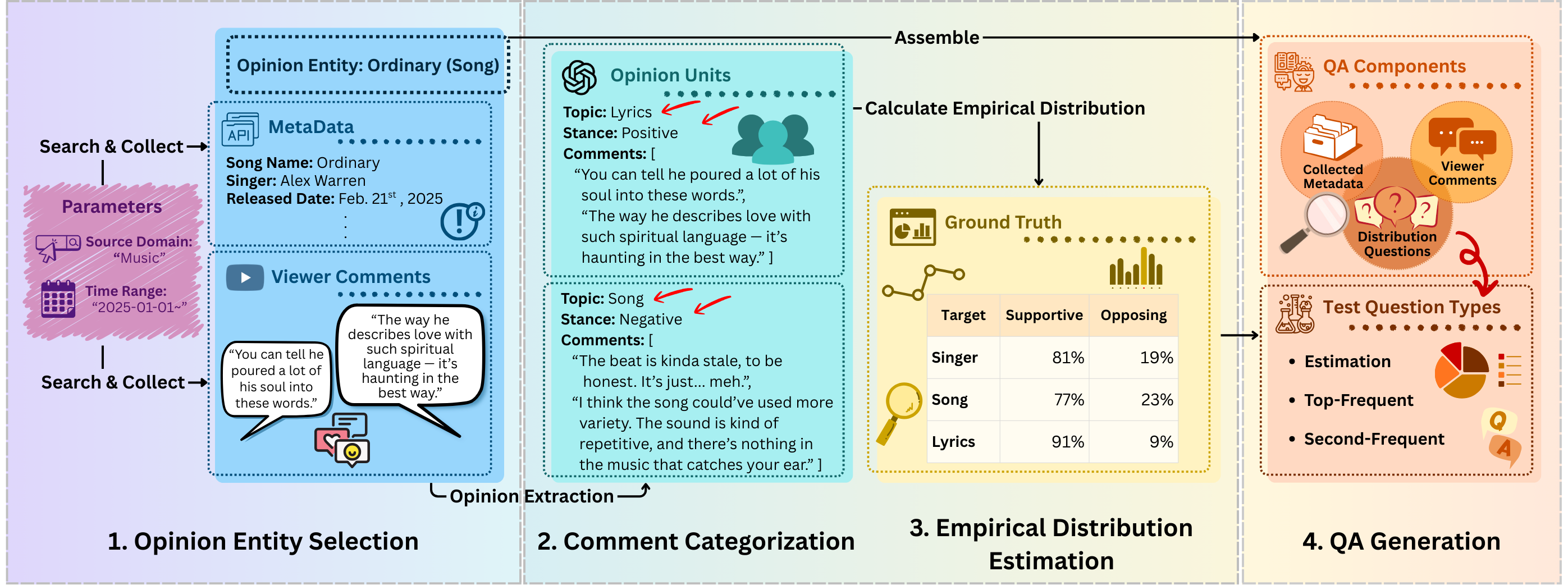}
    \caption{\textbf{\textsc{Text2DistBench} Generation Pipeline.}
    The construction process consists of four stages:
    (1) selecting valid opinion entities;
    (2) annotating comment topic and sentiment;
    (3) estimating empirical opinion distributions; and
    (4) generating distributional reading-comprehension questions.}
    \label{fig:data_gen}
\end{figure*}

\textsc{Text2DistBench} is constructed through a fully automated data generation pipeline designed to support continuous updates.
As illustrated in Figure~\ref{fig:data_gen}, the pipeline consists of four stages:
(1) entity selection,
(2) comment annotation,
(3) empirical distribution estimation, and
(4) QA generation.

\paragraph{Stage 1: Opinion Entity Selection}
To reliably evaluate whether models can infer distributional knowledge from the provided text, it is crucial that target entities are not already familiar to the model. If an entity has been widely discussed during pretraining, the model may have already encountered entity-related comments or human opinions, rather than actually comprehending the provided input, leading to a contaminated evaluation.

To mitigate this issue, we select entities according to two criteria. First, each entity must appear at least six months after the model’s pretraining cutoff date, reducing the likelihood that large-scale public discussions are included in the training data. 
Second, entities must be largely self-contained, such that audience reactions are primarily driven by the entity itself rather than by other similar or previously existing entities.
Based on these criteria, newly released movies and musics are well-suited domains for our benchmark.

In practice, we retrieve candidate entities using domain-specific APIs, including IMDB and TMDB for movies and YouTube Music for songs.
These APIs allow us to filter entities by release time and obtain relevant metadata.
For each entity, we then use the YouTube Data API to retrieve associated videos (e.g., movie trailers or music videos).
Entities with comment volumes below a minimum threshold are discarded.
For the retained entities, viewer comments and metadata are collected.
Additional details of the retrieval and filtering process are provided in Appendix~\ref{appendix:data_gen:vid_match}.

% An LLM (\textsc{GPT-4o-mini}) serves as a verifier to filter out unrelated or noisy retrieval results. Entities with comment volumes below a minimum threshold are discarded.

\paragraph{Stage 2: Comment Annotation}
% For each selected entity, viewer comments are automatically annotated with two attributes: \textit{topic} and \textit{sentiment}. 
% Following prior work showing that LLMs can outperform well-trained human annotators on common text annotation tasks (e.g., tweet topic, sentiment labeling)~\citep{Gilardi_2023},  multiple state-of-the-art LLMs independently assign a \emph{(\text{{topic}}, \text{sentiment})} label to each comment.
% We further apply majority voting across LLM outputs to mitigate model-specific biases and improve annotation robustness~\citep{farr2024llmchainensemblesscalable, qiu2025labelingfreetextdatausing}. 
For each selected entity, viewer comments are automatically annotated with two attributes: \textit{topic} and \textit{sentiment}. 
Motivated by prior work showing that LLMs can match or even outperform human annotators on common text classification tasks~\citep{Gilardi_2023}, we employ multiple state-of-the-art LLMs to independently assign a \emph{(topic, sentiment)} label to each comment. 
We then apply majority voting across model outputs to mitigate model-specific biases and improve annotation robustness~\citep{farr2024llmchainensemblesscalable, qiu2025labelingfreetextdatausing}. 
Comments for which no majority is reached, or that are identified as expressing multiple topics, are discarded.

To assess annotation quality, we conduct human verification on all comments in the dataset (Section~\ref{sec:experiment_setting}). The automated annotations achieve 95\% precision against human judgments, indicating high annotation reliability. Details of the annotation process, prompt design, and human evaluation are provided in Appendix~\ref{appendix:data_gen:comment_label} and \ref{appendix:gen_prompt}.

\paragraph{Stage 3: Empirical Distribution Estimation} 
Following annotation, each entity is associated with a collection of comments, each assigned a \textit{(topic, sentiment)} label.
Treating each comment as one observation, we compute an empirical joint distribution by counting label frequencies and normalizing over all comments.
From this joint distribution, we derive the corresponding marginal and conditional distributions, which serve as ground-truth statistics for distributional question answering.

\paragraph{Stage 4: QA Generation}
Finally, we generate distributional reading-comprehension questions using the templates described in Section~\ref{sec:dataset}.
Each benchmark instance consists of entity metadata, the associated viewer comments, and a set of question–answer pairs, where each answer is computed directly from the empirical distributions.
Examples are provided in Appendix~\ref{appendix:prompt}.

\begin{table*}[t]
\centering
\small
\setlength{\tabcolsep}{5pt}
\begin{tabular}{lcccc|cccc|cccc}
\toprule
& \multicolumn{4}{c}{\textbf{Estimation (1-TVD) $\uparrow$}} 
& \multicolumn{4}{c}{\textbf{Most Frequent (ACC) $\uparrow$}} 
& \multicolumn{4}{c}{\textbf{Second Frequent (ACC) $\uparrow$}} \\
\cmidrule(lr){2-5}\cmidrule(lr){6-9}\cmidrule(lr){10-13}
Model 
& M & C & J & Avg 
& M & C & J & Avg 
& M & C & J & Avg \\
\midrule
GPT-5.1           
& 0.930 & 0.865 & 0.847 & \textbf{0.881} 
& 0.861 & 0.877 & 0.778 & \textbf{0.839} 
& 0.834 & 0.805 & 0.556 & 0.732 \\

Gemini-2.5-Pro    
& 0.903 & 0.869 & 0.805 & 0.859 
& 0.805 & 0.886 & 0.722 & 0.804 
& 0.778 & 0.835 & 0.667 & \textbf{0.760} \\

Grok-4-Fast       
& 0.911 & 0.856 & 0.838 & 0.868 
& 0.889 & 0.858 & 0.667 & 0.805 
& 0.833 & 0.728 & 0.611 & 0.724 \\

Claude-Sonnet-4.5 
& 0.907 & 0.818 & 0.828 & 0.851 
& 0.833 & 0.809 & 0.611 & 0.751 
& 0.805 & 0.728 & 0.500 & 0.678 \\

Llama-3.3-70B     
& 0.855 & 0.836 & 0.744 & 0.812 
& 0.778 & 0.819 & 0.667 & 0.755 
& 0.805 & 0.728 & 0.278 & 0.604 \\

Qwen3-32B         
& 0.701 & 0.747 & 0.619 & 0.689 
& 0.777 & 0.743 & 0.500 & 0.673 
& 0.555 & 0.495 & 0.167 & 0.406 \\
\midrule
Random Baseline
& 0.651 & 0.594 & 0.485 & 0.577 
& 0.375 & 0.435 & 0.132 & 0.314 
& 0.375 & 0.444 & 0.125 & 0.315 \\
\bottomrule
\end{tabular}
% \caption{\textbf{\textsc{Text2DistBench} Results}. M/C/J denote Marginals, Conditionals, and Joint.}
\caption{\textbf{\textsc{Text2DistBench} Results}
We report performance across distribution
(\textbf{M}arginals, \textbf{C}onditionals, \textbf{J}oint) and task types.
Random baselines are computed from 50 runs of random guessing for Most/Second Frequent questions and from a uniform distribution for Estimation questions. Higher values indicate better performance.}
\label{tab:main_results}
\end{table*}

\paragraph{Benchmark Maintenance}
\textsc{Text2DistBench} supports easy maintenance. 
By periodically retrieving newly released entities and re-running the automated generation pipeline, the benchmark can be continuously updated with new content.
The entire process requires no human annotation, enabling up-to-date evaluation for newly released LLMs.

% \section{Experimental Setup}
% \label{sec:experiment_setting}
% Using \textsc{Text2DistBench}, we evaluate a range of state-of-the-art LLMs.
% modified
% \section{Demonstrating Text2DistBench}
\section{Experimental Setup}
\label{sec:experiment_setting}
% In this section, we demonstrate how \textsc{Text2DistBench} can be exploited to evaluate the quality of existing LLMs. 
% In this section, we describe the experimental setting of \textsc{Text2DistBench}.

\paragraph{Models}  
We evaluate both closed-source and open-source LLMs, including 
\textsc{GPT-5.1}~\footnote{\url{https://platform.openai.com/docs/models/gpt-5.1}},  
\textsc{Gemini-2.5-Pro}~\footnote{\url{https://aistudio.google.com/app/prompts/new_chat?model=gemini-2.5-pro}}, 
\textsc{Grok-4-Fast}~\footnote{\url{https://x.ai/news/grok-4-fast}}, 
\textsc{Claude-Sonnet-4.5}~\footnote{\url{https://platform.claude.com/docs/en/about-claude/models/overview}}, 
\textsc{Qwen3-32B}~\footnote{\url{https://huggingface.co/Qwen/Qwen3-32B}}, and 
\textsc{Llama3.3-70B}~\footnote{\url{https://huggingface.co/meta-llama/Llama-3.3-70B-Instruct}}. These models vary in pretraining data, parameter scale, and alignment strategies, providing a diverse testbed for evaluation.

\paragraph{Benchmark Configuration} Since \textsc{Text2Dist-}\textsc{Bench} is continuously updated, we construct a dataset using the pipeline in Section~\ref{sec:dataset:construction}.
The dataset consists of 20 movie and music entities released between July and
October 2025,\footnote{The most recent model evaluated, \textsc{Gemini-2.5-Pro},
has a knowledge cutoff date of 2025-01.}
Under this configuration, each task type yields 40 marginal, 120 conditional,
and 20 joint questions, resulting in a total of 540 evaluation questions across
the three tasks. For each entity, we randomly sample 50 comments from the associated videos to balance question token length. Dataset statistics and construction details are reported in Appendix~\ref{appendix:data_stats}.

\paragraph{Inference Configuration} 
We evaluate all models using fixed zero-shot prompt templates.
Each prompt includes definitions of sentiment and topic, the entity metadata, a set of associated comments, and a distributional question. Full prompt templates are provided in Appendix~\ref{appendix:prompt}.
All models are evaluated in text-only mode, without tool calling or retrieval.
Each question is sampled once using the model’s default decoding parameters, including temperature and maximum token limits. 

% modified
% As contemporary LLMs increasingly allocate more reasoning effort within a single response rather than relying on multiple independent samples and consistency-based aggregation, we generate one sample per instance. This setting reflects common inference practice while controlling computational cost.

\section{Results \& Analysis}
\label{sec:experiment_result}
Building on the setup described in Section~\ref{sec:experiment_setting}, we first present the main results, followed by additional analyses on model behavior.

\begin{figure*}[t]
\centering
\includegraphics[width=\linewidth]{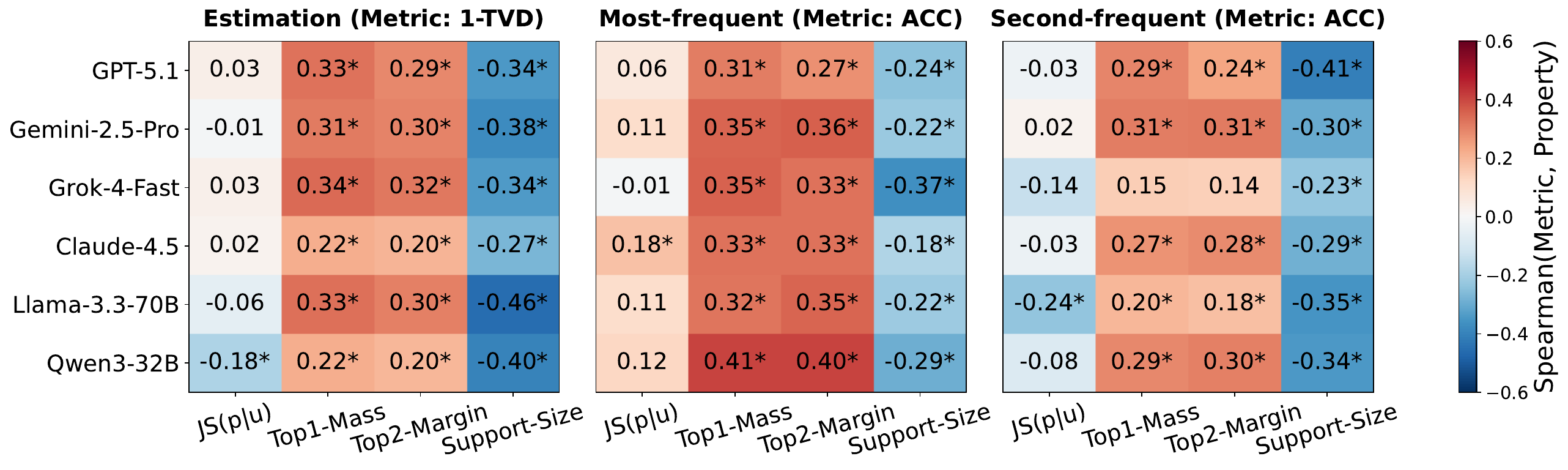}
\caption{\textbf{Sensitivity to target distribution characteristics.} Each cell reports the Spearman correlation between model performance and distribution statistics. (*) indicates statistical significance ($pvalue < 0.05$).}
\label{fig:dist_property}
\end{figure*}

% \begin{figure*}[t]
% \centering
% \includegraphics[width=\linewidth]{latex/figure/domain_effect.pdf}
% \caption{\textbf{Domain gap on model performance.}
% Average performance difference between questions of different source domains (Music -- Movie). Negative(Positive) values indicate better performance on movie(music) reviews.}
% \label{fig:domain}
% \end{figure*}

\begin{figure*}[t]
\centering
\includegraphics[width=\linewidth]{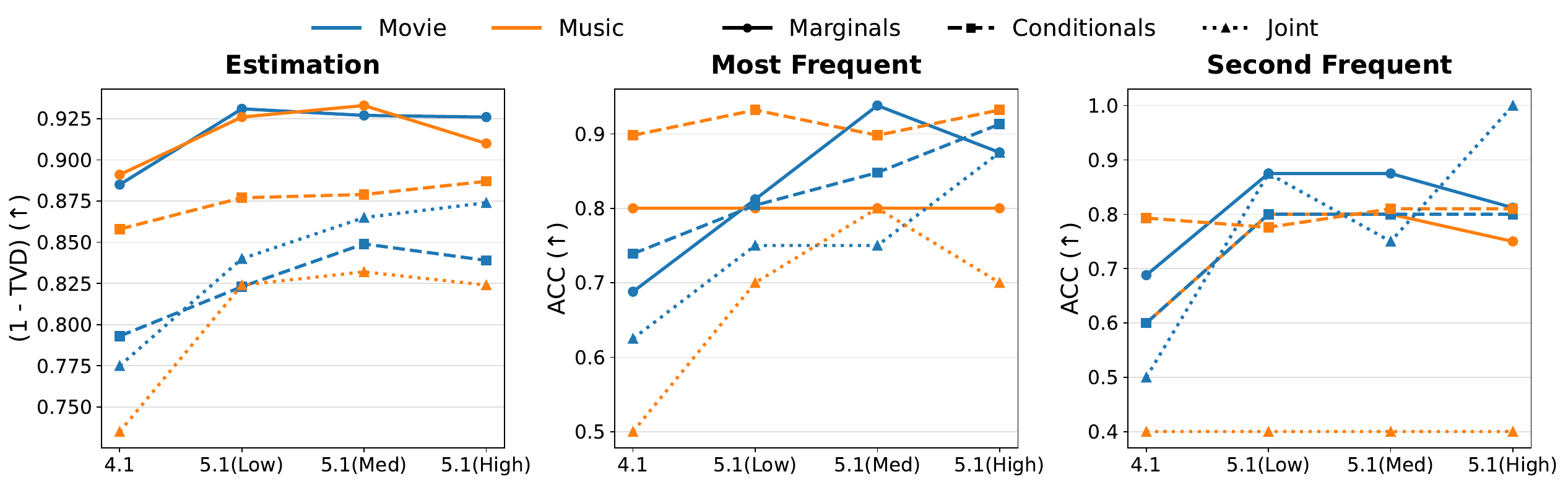}
\caption{\textbf{Effect of scaling reasoning effort.}
Performance of GPT-5.1 under different reasoning-effort settings, with GPT-4.1 shown as a non-reasoning baseline for comparison.}
\label{fig:reason}
\end{figure*}

\begin{figure*}[t]
\centering
\includegraphics[width=\linewidth]{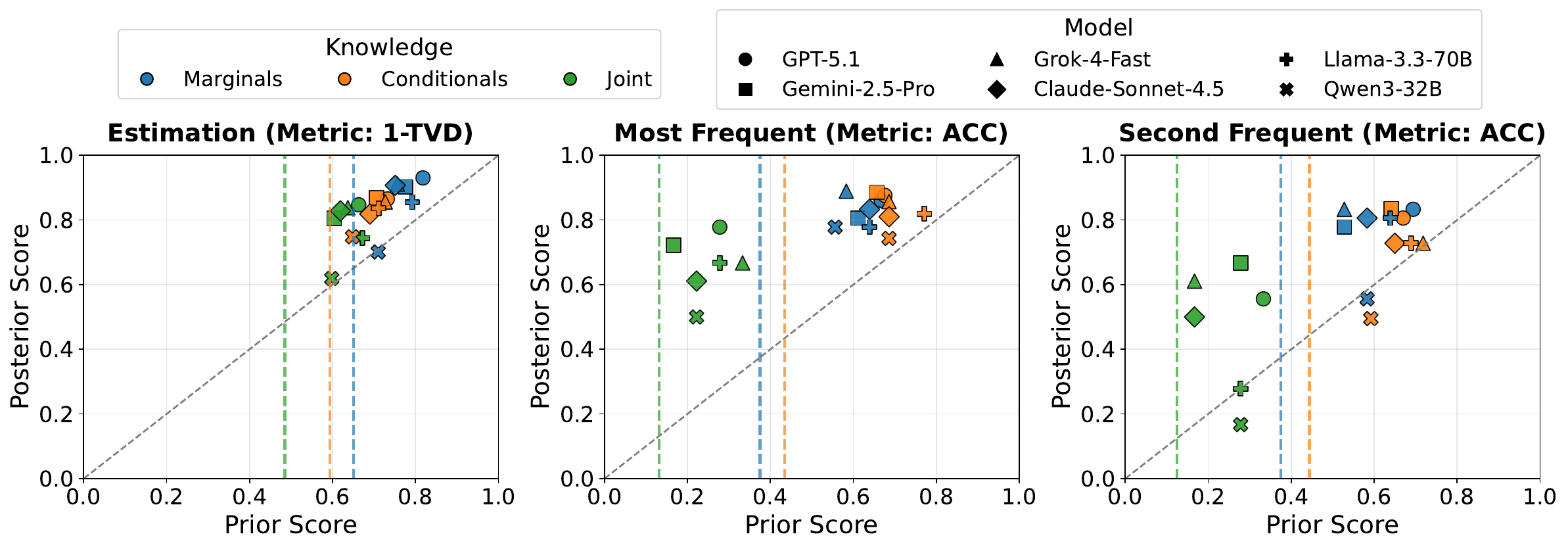}
\caption{\textbf{Prior belief from factual information.}
Each subplot compares prior (metadata-only) and posterior (with comments) performance for one task. The diagonal denotes equal prior and posterior performance, with points above it indicating improvements after observing viewer comments. 
Colored vertical dashed lines indicate random baselines for each distribution type.
% Colored horizontal dashed lines indicate random baselines for each distribution type.
}
\label{fig:prior}
\end{figure*}

\subsection{Overall Performance and Ranking}
\label{sec:experiment_result:overall}
Table~\ref{tab:main_results} summarizes model performance on \textsc{Text2DistBench} across different distribution and task types. Across all settings, every model substantially outperforms the random baseline, indicating that current LLMs can understand non-trivial distributional information from text. 

% \paragraph{Model Ranking across Tasks.}
Model rankings differ across task types. 
For estimation task, \textsc{GPT-5.1} achieves the highest average
performance, with \textsc{Gemini-2.5-Pro} and \textsc{Grok-4-Fast} closely
following, and a similar ordering is observed for most-frequent task.
In contrast, second-frequent task favors \textsc{Gemini-2.5-Pro}, with
\textsc{GPT-5.1} and \textsc{Grok-4-Fast} ranking slightly lower.
Mid-tier models such as \textsc{Claude-Sonnet-4.5} and \textsc{Llama-3.3-70B} remain competitive on estimation tasks but exhibit larger drops on most/second frequent questions, while \textsc{Qwen3-32B} consistently ranks lowest despite outperforming the random baseline.

% \paragraph{Structural Difficulty of Distribution Types.}
% Across all tasks, a consistent difficulty ordering emerges across distribution types. Marginal questions are the easiest for models, followed by conditional questions, while joint distribution questions are the most challenging.
% This trend holds for estimation task and is especially pronounced for most-frequent and second-frequent questions, where performance drops sharply on joint distributions.
Across all tasks, a consistent difficulty ordering emerges across distribution types.
Marginal questions are the easiest for models, followed by conditional questions, while joint distribution questions are the most challenging.
This trend holds for estimation tasks and is especially pronounced for most/second frequent questions, where performance drops sharply on joint distributions.

Overall, while several models consistently form a top-performing group, their performance varies across distribution and task types. 
This implies that, in practice, model selection should depend on the specific information the user is interested in.

% modified
% Among the evaluated models, \textsc{GPT-5.1} achieves the highest average performance across all tasks, with \textsc{Gemini-2.5-Pro} and
% \textsc{Grok-4-Fast} closely following. In contrast, \textsc{Qwen3-32B} consistently yields the lowest scores.Across all models, we observe a consistent difficulty ordering among \emph{distribution types}: questions over marginal distributions are the easiest, followed by conditional distributions, while joint distributions are the most challenging. Across tasks, models generally perform worse on second-frequent than on estimation and most-frequent. Despite differences in absolute performance, relative model rankings remain largely stable across distribution and tasks.

\subsection{Target Distribution Characteristics}
\label{sec:experiment_result:distribution}
Figure~\ref{fig:dist_property} analyzes how model performance correlates with
intrinsic properties of the target distributions.
We characterize each ground-truth distribution using four statistics:
(1) \emph{JS divergence from the uniform distribution}, which measures the degree of non-uniformity;
(2) \emph{Top-1 probability mass}, which captures the dominance of the most frequent category;
(3) \emph{Top-2 margin}, defined as the probability difference between the most and second most frequent categories, reflecting how sharply peaked the distribution is; and
(4) \emph{Support size}, the number of categories with non-zero probability mass.

Across models and tasks, we observe consistent trends. 
Performance exhibits positive rank correlations with both Top-1 probability mass and Top-2 margin, indicating that distributions with stronger dominance and clearer separation between leading categories are easier for models to handle.
In contrast, support size shows a negative correlation with performance, suggesting that distributions spread across more categories pose greater difficulty.
Finally, performance shows little systematic correlation with JS divergence from the uniform distribution, indicating that overall non-uniformity alone is not a strong predictor of model success.

% \subsection{Text Source Domain}
% \label{sec:experiment_result:domain}
% Figure~\ref{fig:domain} analyzes the average performance gap between the movie and music domain questions. For estimation task, domain gaps are consistently small, with performance differences remaining close to zero across all models.
% In contrast, domain gaps are more pronounced for most-frequent task. Both the magnitude and direction of the gap vary across models: \textsc{Grok-4-Fast} shows a strong advantage on movie reviews, whereas \textsc{Qwen3-32B} performs better on music reviews, with differences reaching approximately 15\%. Other models exhibit smaller but noticeable domain preferences. The domain gap is strongest and most consistent across models for second-frequent task. Most models perform better on movie reviews than on music reviews, with performance gaps ranging from 10\% to 15\%. 

\subsection{Scaling Reasoning Effort}
\label{sec:experiment_result:reason}
As an additional analysis, we examine whether increasing reasoning effort improves model performance. We use \textsc{GPT-5.1} as the test model and vary the reasoning effort from low to high. Figure~\ref{fig:reason} shows the performance of \textsc{GPT-5.1} under different reasoning-effort settings, with \textsc{GPT-4.1}, a strong non-reasoning model from the same family, included as a comparison baseline.

Overall, increasing reasoning effort leads to consistent performance improvements, with the largest gains observed for joint distribution questions. 
Across task types, estimation and most-frequent questions benefit more from additional reasoning effort than second-frequent questions.
Performance improves as reasoning effort increases up to a medium level, after which gains diminish and may slightly degrade, indicating diminishing returns from further increases in reasoning effort.

\subsection{Prior Belief from Factual Information}
\label{sec:experiment_result:prior}
Beyond the main experiments, we consider an extreme setting in which viewer comments are removed from the input and only entity metadata is provided. This metadata-only setting mirrors a common human behavior of forming initial population-level beliefs from factual information alone, without observing individual responses.

\begin{table}
\small
\centering
\renewcommand{\arraystretch}{1.2}
\begin{tabular}{lccc}
\hline
\textbf{Model} & \makecell[c]{\textbf{Esti-} \\ \textbf{mation}} & \makecell[c]{\textbf{Most} \\ \textbf{Frequent}} & \makecell[c]{\textbf{Second} \\ \textbf{Frequent}} \\ \hline
GPT-5.1           & 0.142  & 0.220* & 0.246* \\
Gemini-2.5-Pro    & 0.215* & 0.221* & 0.316* \\
Grok-4-Fast       & 0.186* & 0.341* & 0.181* \\
Claude-Sonnet-4.5 & 0.069  & 0.274* & 0.158* \\
Llama-3.3-70B     & 0.219* & 0.154*  & 0.122  \\
Qwen3-32B         & 0.255* & 0.304* & 0.077  \\ \hline
\end{tabular}
\caption{Spearman correlation between model prior and posterior performance. (*) indicates 
statistical significance 
$pvalue < 0.05$.}
\label{tab:correlation_results}
\end{table}

Figure~\ref{fig:prior} compares metadata-only (prior) and full-context (posterior) performance across the three task types. Across all models, prior predictions consistently outperform random baselines, indicating that models can leverage factual metadata to form non-trivial distributional hypotheses even for previously unseen entities.
Importantly, incorporating viewer comments leads to consistent performance improvements, with most points lying above the diagonal, including cases where prior predictions (e.g., joint distribution) are weak.

Table~\ref{tab:correlation_results} further shows that prior and posterior performance are positively correlated for most models and tasks.
This suggests that models with stronger prior distributional beliefs tend to benefit more after observing comments.
Together, these results demonstrate that models actively integrate textual evidence to refine their initial beliefs, highlighting the value of \textsc{Text2DistBench} for evaluating text-conditioned distributional understanding.

% This demonstrates that models actively integrate textual evidence to refine their initial beliefs, highlighting the value of \textsc{Text2DistBench} for evaluating text-conditioned distributional understanding.

\section{Conclusion}
\label{sec:conclusion}
In this work, we introduced \textsc{Text2DistBench}, an automated and continuously
updatable benchmark for evaluating distributional reading comprehension in LLMs.
Our benchmark probes models’ ability to infer distribution knowledge from text, spanning marginal, conditional, and joint distributions under multiple question types. 
Our empirical results show that while LLMs substantially outperform random baselines, their performance varies widely across different distribution settings.
We also find that LLMs can form informative prior beliefs for unseen entities from factual information alone, and that models with stronger priors tend to benefit more from incorporating viewer comments.
Together, these findings highlight both the capabilities and limitations of current LLMs in distributional reading comprehension, and demonstrate the value of \textsc{Text2DistBench} as a practical and scalable testbed for future research.

% In this work, we introduced \textsc{Text2DistBench}, an automated and continuously
% updatable benchmark for evaluating distributional reading comprehension in LLMs.
% Our benchmark probes models’ ability to infer population-level distributions that are
% expressed in text, spanning marginal, conditional,
% and joint distributions under multiple question formats.
% Through extensive evaluation of LLMs, we highlight both the
% current capabilities and remaining limitations of LLMs in understanding
% distributional information from text, and demonstrate the value of \textsc{Text2DistBench} as a practical and scalable testbed for future research.

\section*{Limitations}
\label{sec:limit}
While \textsc{Text2DistBench} provides an automated and continuously updatable benchmark for systematically evaluating distributional reading comprehension, several limitations remain that could be addressed in future work.
% \textbf{Query Variety.}
Currently, the benchmark defines three tasks.
However, in real-world scenarios, humans may ask a wider range of questions about distributional information expressed in text, such as threshold-based or comparative queries.
Extending the framework to support a richer set of query types would enable a more comprehensive evaluation of distributional understanding.
% \textbf{Domain Coverage.} 
Moreover, expanding the benchmark to more domains could further improve its coverage and provide a more diverse testbed for evaluation.

\section*{Ethical Considerations}
\label{sec:ethic}
\textsc{Text2DistBench} is constructed using publicly available APIs, including
IMDB/TMDB for movie metadata and YouTube Data API for music metadata and viewer comments.
While these sources may occasionally contain inaccuracies due to human error or
reporting delays, they are widely recognized and provide verifiable records of
real-world content.
All data are derived from post-release or officially published materials, and no
private, sensitive, or personally identifiable information is included in the
benchmark.

\section*{Use of Ai Assistants}
\label{sec:ai}
In this work, we leveraged large language models (LLMs) to assist research in two ways.
First, multiple LLMs were employed as part of the automated benchmark generation
pipeline, including comment annotation and relevant video verification.
Second, an AI assistant (OpenAI GPT-5.2) was used for minor writing support, such as grammar correction and improving manuscript clarity. All AI-assisted contents were carefully reviewed by the authors to ensure factual accuracy and consistency with the authors’ original intent.

\section*{Acknowledgment}
\label{sec:ack}
This material is based upon work supported by National Science and Technology Council, ROC under grant number 114-2221-E-002-134-MY3 and 113-2628-E-001-003-MY4, NTU AI Center of Research Excellence within Taiwan Centers of Excellence in Artificial Intelligence, and by National Taiwan University and Academia Sinica Innovative Joint Program, under grant AS-NTU-114-06.

\bibliography{latex/custom}

\appendix

\appendix
% switch to single-column mode for appendix
\onecolumn 

\section{Benchmark Statistics}
\label{appendix:data_stats}

Following the experimental setup described in Section~\ref{sec:experiment_setting}, we construct \textsc{Text2DistBench} using entities released between July and October 2025. The benchmark contains 20 entities in total, evenly split across two domains (10 movies and 10 music entities). Each entity is annotated with two attributes: \emph{sentiment} ($|S|=2$) and \emph{topic} ($|T|=4$), which together define the underlying distributions.
For each entity, we generate reading comprehension questions over three distribution types: marginals ($P(S)$ and $P(T)$), conditionals ($P(S\!\mid\!T)$ and $P(T\!\mid\!S)$), and the joint distribution ($P(S,T)$). Each question is instantiated under three \emph{task types}: Estimation, Most-Frequent, and Second-Frequent.
For a single task, each entity is associated with a fixed set of questions:
(1) 2 marginal questions ($P(S)$ and $P(T)$);
(2) 6 conditional questions (4 for $P(S\!\mid\!T)$ and 2 for $P(T\!\mid\!S)$); and
(3) 1 joint question ($P(S,T)$). 
This results in 9 distributional questions per entity per task type.
Across the three tasks, each entity therefore yields 27 evaluation questions.
Table~\ref{tab:data_stats} summarizes the number of questions per task and the total number of evaluation instances across all tasks.

\begin{table}[htbp]
\small
\centering
\begin{tabular}{lcccc|c}
\toprule
\textbf{Entities} 
& \textbf{\#Marginals} 
& \textbf{\#Conditionals} 
& \textbf{\#Joints} 
& \textbf{\#Total (per Task)} 
& \textbf{\#Total (3 Tasks)} \\
\midrule
20 (10 Movie + 10 Music) 
& 40 
& 120 
& 20 
& $180$ 
& 540 \\
\bottomrule
\end{tabular}
\caption{\textbf{Benchmark Statistics Summary.}}
\label{tab:data_stats}
\end{table}

\section{Data Generation}
\label{appendix:data_gen}
This section describes the \emph{Opinion Entity Selection} and
\emph{Comment Annotation} stages of the automated generation pipeline used to
construct \textsc{Text2DistBench}.

\subsection{Opinion Entity Selection}
\label{appendix:data_gen:vid_match}
We retrieve candidate opinion entities using domain-specific APIs, including
IMDB and TMDB for movies and YouTube Music for songs. These APIs allow us to filter entities by release time and obtain relevant metadata.
For each entity, we then use the YouTube Data API to retrieve associated videos,
such as movie trailers or music videos.
Because the YouTube Data API relies on text-based matching, some retrieved videos may be unrelated to the target entity. To address this issue, we employ an LLM (\textsc{GPT-5.1-mini}) to filter out irrelevant or noisy video results.
The prompt template used for this verification step is shown in Appendix~\ref{appendix:gen_prompt}. Entities with comment volumes below a minimum threshold are discarded. For the remaining entities, we collect viewer comments and associated metadata.

\subsection{Comment Annotation}
\label{appendix:data_gen:comment_label}
\paragraph{Method} 
We use three API-based LLMs (\textsc{GPT-5.1-mini}, \textsc{Gemini-2.5-Flash},
and \textsc{Grok-4-Fast}) to annotate viewer comments.
For each comment, models are prompted to assign both a \emph{topic} and a
\emph{sentiment}.
Tables~\ref{tab:movie} and~\ref{tab:music} list the topic definitions provided
to the models for the movie and music domains, respectively.
All models are prompted using the template shown in Appendix~\ref{appendix:gen_prompt}.
When model predictions disagree, we apply a majority-voting scheme to determine
the final label. If no majority is reached, or if a model indicates that a comment expresses multiple topics, the comment is discarded.

\paragraph{Annotation Quality}
To assess annotation quality, we conduct human verification on the test set
described in Section~\ref{sec:experiment_setting}, which consists of 20 entities
with 50 comments each, totaling 1{,}000 comments.
Human annotators are provided with the same annotation instructions as the
models, including label definitions, entity metadata, and viewer comments.
We recruit three annotators (two undergraduate students and one graduate
student), and determine the final human label for each comment by majority vote.
Annotation precision is then computed by comparing the automated labels against
the human annotations. 

Based on the evaluation, with n=1{,}000 and an observed precision of 0.95, the sampling distribution follows a Binomial distribution, which, according to the Central Limit Theorem, yields a 95\% confidence interval of [0.936, 0.964] with a margin error of ±1.4\%. This narrow interval indicates that the estimated precision is statistically stable, suggesting that our annotation pipeline produces results that are largely consistent with human judgments.

\begin{table*}[h!]
\centering
\small
\begin{tabular}{p{2cm} p{10cm}}
\toprule
\textbf{Attribute} & \textbf{Description} \\
\midrule
Actor & Performance, emotion, chemistry, casting quality. \\
Storyline & Plot, pacing, themes, character arcs, dialogue. \\
Visual & Cinematography, lighting, color, design, VFX, costumes. \\
Audio & Soundtrack, score, songs, sound effects, audio mix. \\
\bottomrule
\end{tabular}
\caption{Movie Comment Topics}
\label{tab:movie}
\end{table*}

\begin{table*}[h!]
\small
\centering
\begin{tabular}{p{2cm} p{10cm}}
\toprule
\textbf{Attribute} & \textbf{Description} \\
\midrule
Song & Melody, harmony, rhythm, structure, arrangement. \\
Singer & Tone, technique, emotion, appearance. \\
Lyrics & Themes, message, storytelling, rhymes. \\
Visual & Cinematography, lighting, concept, choreography, VFX. \\
\bottomrule
\end{tabular}
\caption{Music Comment Topics}
\label{tab:music}
\end{table*}

\section{Data Generation Prompt Examples (Movie)}
\label{appendix:gen_prompt}

\begin{tcolorbox}[promptbox, title={Movie Metadata}]
    \texttt{- Movie Title: \{title\}} \\ 
    \texttt{- Release Date: \{date\}} \\ 
    \texttt{- Cast: \{casts\}} \\ 
    \texttt{- Summary: \{summary\}} \\ 
    \texttt{- Synopsis: \{synopsis\}} 
\end{tcolorbox}

\begin{tcolorbox}[promptbox, title={Relevant Video Verification Prompt}]
\texttt{You searched YouTube for: \{query\}  \\
Following is the retrieved video information:  \\
Title: \{title\}  \\
Description: \{description\}   \\
\\
Determine if this video matches the intended search query and satisfies the information need. \\
If yes, return True. If not, return False.
}
\end{tcolorbox}

\begin{tcolorbox}[promptbox, title={Comment Topic Annotation Prompt}]
\texttt{You are analyzing public reactions to a movie by assigning each viewer comment to one or more attributes (multi-label). \\
\\
Attributes (use EXACTLY these keys; prefer the single most dominant attribute unless the comment clearly discusses multiple): \\
- Actor : Comments about the actors’ performances: delivery, emotion, chemistry, casting. \\
- Storyline : Comments about the movie plot, narrative, themes, pacing, character arcs, or dialogue. \\
- Visual : Comments about cinematography, animation, lighting, color, production design, costumes, or visual effects. \\
- Audio : Comments about the soundtrack, score, songs, sound effects, or audio mix. \\
- Other : Use only if none of the above clearly fit (off-topic, spam, unclear). \\
\\
Movie Information: \\ \{meta\_data\} \\
\\
YouTube Viewer Comments (0-based indexing, e.g., 0, 1, 2, ...): 
\\
\{comments\} \\
\\
Instructions: \\
1) MULTI-LABEL is allowed, but if uncertain choose the single most dominant attribute. \\
2) Use 0-based indices exactly as shown; do not invent indices. \\
3) If a comment does not clearly fit any attribute, include it under "Other". \\
4) Return ONLY the JSON object. \\
5) The JSON must be a single object whose keys are EXACTLY the attributes below and whose values are lists of integer indices. \\
6) Do not add, rename, or remove keys. \\
\\
Output JSON: \\
\{"Actor": [], "Storyline": [], "Visual": [], "Audio": [], "Other": []\}
}
\end{tcolorbox}

\begin{tcolorbox}[promptbox, title={Comment Sentiment Annotation Prompt}]
\texttt{Classify the sentiment expressed in YouTube comments toward the movie. \\
\\
Movie Information (for information reference only): \\
\{meta\_data\} \\
\\
YouTube Viewer Comments (0-based indexing, e.g., 0, 1, 2, ...): \\
\{comments\} \\
\\
Labels (choose exactly one per comment): \\
- support : praise / approval / positive attitude \\
- oppose  : criticism / disapproval / negative attitude \\
\\
Rules: \\
1) Focus on the overall tone or attitude of the comment toward the movie. \\
2) Use movie information only to resolve references (e.g., who/what "he", "it", or "this scene" refers to), not to guess sentiment. \\
3) Consider emojis, slang, irony/sarcasm (e.g., quotes, “/s”, exaggeration, laugh reactions). \\
4) Use 0-based indices exactly as shown; do not invent or skip indices. \\
5) Each index must appear in EXACTLY ONE list (support OR oppose). \\
\\
Output JSON: \\
\{"support": [], "oppose": []\}
}
\end{tcolorbox}

\section{QA Template Examples (Movie)}
\label{appendix:prompt}

\begin{tcolorbox}[promptbox, title={System Template}]
\texttt{You will be given information about a movie, followed by a collection of viewer comments. \\
Each comment reflects what a viewer thinks about the movie and focuses on a particular aspect while expressing a certain attitude. \\
\\
When reading the comments, keep in mind two dimensions: \\
- Sentiment: whether the comment expresses a positive or negative attitude. \\
- Topic: which aspect of the movie the comment mainly talks about. \\
\\
Sentiment categories: \\
- positive: expressing approval, enjoyment, or praise. \\
- negative: expressing criticism, dissatisfaction, or disappointment. \\
\\
Topic categories: \\
- Actor: acting performance, casting, chemistry, emotional expression. \\
- Storyline: plot, narrative, pacing, themes, dialogue, character development. \\
- Visual: cinematography, animation, lighting, color, visual effects, production design. \\
- Audio: soundtrack, music, sound effects, or audio quality. \\
\\
Read the following movie information and the viewer comments carefully. \\
Then answer the question based on this information. \\
\\
Movie Information: \{meta\_data\} \\
Viewer Comments: \{comments\}
}
\end{tcolorbox}

\begin{tcolorbox}[promptbox, title={$P(S)$ Estimation}]
\texttt{Question: \\
Read the viewer comments.  \\
How do the expressed opinions break down in terms of overall attitude toward the movie? \\
Summarize how common each sentiment is among the comments. \\
Ensure the percentages sum to 100 and use integers only. \\
\\
Output your answer in the following JSON format: \\
\{ "positive": "<int>\%", "negative": "<int>\%"\} \\
Answer:
}
\end{tcolorbox}

\begin{tcolorbox}[promptbox, title={$P(T)$ Estimation}]
\texttt{Question: \\
Read the viewer comments.  \\
What aspects of the movie do viewers talk about? \\
Summarize how frequently each aspect appears in the comments. \\
Ensure the percentages sum to 100 and use integers only. \\
\\
Output your answer in the following JSON format: \\
\{"Actor": "<int>\%", "Storyline": "<int>\%", "Visual": "<int>\%", "Audio": "<int>\%"\} \\
Answer:
}
\end{tcolorbox}

\begin{tcolorbox}[promptbox, title={$P(S,T)$ Estimation}]
\texttt{Question: \\
Read the viewer comments.  \\
Consider both which aspect is being discussed and whether the expressed opinion is positive or negative. \\
Summarize how these combinations appear in the comments. \\
Ensure the percentages sum to 100 and use integers only. \\
\\
Output your answer in the following JSON format: \\
\{ \\
\ \ \ \ "percentages": \{ \\
\ \ \ \ \ \ \ \ "(Actor,positive)": "<int>\%", \\
\ \ \ \ \ \ \ \ "(Actor,negative)": "<int>\%", \\
\ \ \ \ \ \ \ \ "(Storyline,positive)": "<int>\%", \\
\ \ \ \ \ \ \ \ "(Storyline,negative)": "<int>\%", \\
\ \ \ \ \ \ \ \ "(Visual,positive)": "<int>\%", \\
\ \ \ \ \ \ \ \ "(Visual,negative)": "<int>\%", \\
\ \ \ \ \ \ \ \ "(Audio,positive)": "<int>\%", \\
\ \ \ \ \ \ \ \ "(Audio,negative)": "<int>\%" \\
\ \ \ \ \} \\
\} \\
Answer:
}
\end{tcolorbox}

\begin{tcolorbox}[promptbox, title={$P(S|T)$ Estimation}]
\texttt{Question: \\
Read the viewer comments.  \\
Focus only on the comments that talk about the \{topic\} aspect of the movie. \\
How are viewers’ attitudes divided? \\
Ensure the percentages sum to 100 and use integers only. \\
\\
Output your answer in the following JSON format: \\
\{"positive": "<int>\%", "negative": "<int>\%" \} \\
Answer:
}
\end{tcolorbox}

\begin{tcolorbox}[promptbox, title={$P(T|S)$ Estimation}]
\texttt{Question: \\
Read the viewer comments.  \\
Focus only on the comments that express a \{sentiment\_label\} attitude toward the movie. \\
Which aspects of the movie do these comments discuss? \\
Ensure the percentages sum to 100 and use integers only. \\
\\
Output your answer in the following JSON format: \\
\{"Actor": "<int>\%", "Storyline": "<int>\%", "Visual": "<int>\%", "Audio": "<int>\%"\} \\
Answer:
}
\end{tcolorbox}

% ---------- MOSTFREQ_S_TEMPLATE ----------
\begin{tcolorbox}[promptbox, title={$P(S)$ Most Frequent}]
\texttt{Question: \\
Read the viewer comments.   \\
What overall attitude do most viewers express? \\
Positive or Negative. \\
Answer:
}
\end{tcolorbox}

% ---------- MOSTFREQ_T_TEMPLATE ----------
\begin{tcolorbox}[promptbox, title={$P(T)$ Most Frequent}]
\texttt{Question: \\
Read the viewer comments.  \\
Which aspect of the movie is discussed most often? \\
Choose from: Actor, Storyline, Visual, or Audio. \\
Answer:
}
\end{tcolorbox}

% ---------- MOSTFREQ_T_S_TEMPLATE ----------
\begin{tcolorbox}[promptbox, title={$P(S,T)$ Most Frequent}]
\texttt{Question: \\
Read the viewer comments.  \\
Considering both (1) which aspect is being talked about and (2) whether the attitude is positive or negative. \\
Which combination appears most often in the comments? \\
Choose one pair from: \\
(Actor,positive), (Actor,negative), \\
(Storyline,positive), (Storyline,negative), \\
(Visual,positive), (Visual,negative), \\
(Audio,positive), (Audio,negative). \\
Answer:
}
\end{tcolorbox}

% ---------- MOSTFREQ_S_cond_T_TEMPLATE ----------
\begin{tcolorbox}[promptbox, title={$P(S|T)$ Most Frequent}]
\texttt{Question: \\
Read the viewer comments. \\
Focus only on the comments that talk about the \{topic\} aspect of the movie. \\
What attitude do viewers most commonly express? \\
Positive or negative.\\
Answer:
}
\end{tcolorbox}

% ---------- MOSTFREQ_T_cond_S_TEMPLATE ----------
\begin{tcolorbox}[promptbox, title={$P(T|S)$ Most Frequent}]
\texttt{Question: \\
Read the viewer comments. \\
Focus only on the comments that express a \{sentiment\_label\} attitude toward the movie. \\
Which aspect of the movie is mentioned most often? \\
Choose from: Actor, Storyline, Visual, or Audio. \\
Answer:
}
\end{tcolorbox}

% ---------- SECONDMOST_S_TEMPLATE ----------
\begin{tcolorbox}[promptbox, title={$P(S)$ Second Most Frequent}]
\texttt{Question: \\
Read the viewer comments.   \\
What overall attitude is the second most commonly expressed by viewers? \\
Positive or Negative\\
Answer:
}
\end{tcolorbox}

% ---------- SECONDMOST_T_TEMPLATE ----------
\begin{tcolorbox}[promptbox, title={$P(T)$ Second Most Frequent}]
\texttt{Question: \\
Read the viewer comments.  \\
Which aspect of the movie is discussed the second most often? \\
Choose from: Actor, Storyline, Visual, or Audio. \\
Answer:
}
\end{tcolorbox}

% ---------- SECONDMOST_T_S_TEMPLATE ----------
\begin{tcolorbox}[promptbox, title={$P(S,T)$ Second Most Frequent}]
\texttt{Question: \\
Read the viewer comments.  \\
Considering both (1) which aspect is being talked about and (2) whether the attitude is positive or negative. \\
Which combination appears the second most often in the comments? \\
Choose one pair from: \\
(Actor,positive), (Actor,negative), \\
(Storyline,positive), (Storyline,negative), \\
(Visual,positive), (Visual,negative), \\
(Audio,positive), (Audio,negative). \\
Answer:
}
\end{tcolorbox}

% ---------- SECONDMOST_S_cond_T_TEMPLATE ----------
\begin{tcolorbox}[promptbox, title={$P(S|T)$ Second Most Frequent}]
\texttt{Question: \\
Read the viewer comments. \\
Focus only on the comments that talk about the \{topic\} aspect of the movie. \\
What attitude is the second most commonly expressed? \\
Positive or Negative\\
Answer:
}
\end{tcolorbox}

% ---------- SECONDMOST_T_cond_S_TEMPLATE ----------
\begin{tcolorbox}[promptbox, title={$P(T|S)$ Second Most Frequent}]
\texttt{Question: \\
Read the viewer comments. \\
Focus only on the comments that express a \{sentiment\_label\} attitude toward the movie. \\
Which aspect of the movie is mentioned the second most often? \\
Choose from: Actor, Storyline, Visual, or Audio. \\
Answer:
}
\end{tcolorbox}

\end{document}